\documentclass[10pt,twocolumn,letterpaper]{article}

\usepackage{iccv}              

\definecolor{iccvblue}{rgb}{0.21,0.49,0.74}
\usepackage[pagebackref,breaklinks,colorlinks,allcolors=iccvblue]{hyperref}

\def\paperID{9127} 
\def\confName{ICCV}
\def\confYear{2025}
\usepackage{multirow}

\title{PHATNet: A Physics-guided Haze Transfer Network for Domain-adaptive Real-world Image Dehazing}

\author{{
{Fu-Jen Tsai{$^{\textcolor{red}{1,2}}$}}\;\;
{Yan-Tsung Peng$^{\textcolor{red}{3}}$}\;\;
{Yen-Yu Lin$^{\textcolor{red}{4}}$}\;\;
{Chia-Wen Lin$^{\textcolor{red}{1}}$}}
\\
{{National Tsing Hua University$^{\textcolor{red}{1}}$}\;\;
{MediaTek$^{\textcolor{red}{2}}$}}
\\
{{National Chengchi University$^{\textcolor{red}{3}}$}\;\;
{National Yang Ming Chiao Tung University$^{\textcolor{red}{4}}$}}
\\
{\tt\small fjtsai@gapp.nthu.edu.tw \;\; ytpeng@cs.nccu.edu.tw \;\; lin@cs.nycu.edu.tw \;\; cwlin@ee.nthu.edu.tw}
}

\begin{document}
\maketitle
\begin{abstract}
Image dehazing aims to remove unwanted hazy artifacts in images. Although previous research has collected paired real-world hazy and haze-free images to improve dehazing models' performance in real-world scenarios, these models often experience significant performance drops when handling unseen real-world hazy images due to limited training data. This issue motivates us to develop a flexible domain adaptation method to enhance dehazing performance during testing. Observing that predicting haze patterns is generally easier than recovering clean content, we propose the Physics-guided Haze Transfer Network (PHATNet) which transfers haze patterns from unseen target domains to source-domain haze-free images, creating domain-specific fine-tuning sets to update dehazing models for effective domain adaptation. Additionally, we introduce a Haze-Transfer-Consistency loss and a Content-Leakage Loss to enhance PHATNet's disentanglement ability. Experimental results demonstrate that PHATNet significantly boosts state-of-the-art dehazing models on benchmark real-world image dehazing datasets. The source code is available at \href{https://github.com/pp00704831/PHATNet}{https://github.com/pp00704831/PHATNet}.

\end{abstract}    
\section{Introduction}
Images captured in hazy conditions often experience significant degradation, including occlusion, color distortion, reduced contrast, etc. These factors severely diminish image clarity and adversely affect downstream computer vision tasks, such as image classification~\cite{Liu_2024_CVPR, rangwani2022closer}, object detection~\cite{hoyer2023mic, Liu_2024_CVPR_2}, and semantic segmentation~\cite{Ma_2022_CVPR, Zhao_2024_CVPR, Yang_2024_CVPR}. Image dehazing seeks to restore high-quality, visually enhanced images from a single hazy input; however, it remains challenging due to its inherently ill-posed nature.

Image dehazing has advanced considerably in recent years, with numerous methods developed using various deep neural network architectures, including Convolutional Neural Networks (CNNs)~\cite{cui2023focal, Wu_2021_CVPR, Zheng_2021_CVPR, Hardgan, FSDGN, qin2020ffa, shen2023mitnet, TANet} and Transformers~\cite{guo2022dehamer, song2023vision}, to improve dehazing performance. However, due to the scarcity of real-world clean-hazy image pairs for training, these methods often rely on synthetic hazy image datasets~\cite{Resides} generated through the following Atmospheric Scattering Model (ASM):    
\begin{equation}
\begin{gathered}
    I(x) = J(x) \times t(x)+A \times (1-t(x))
    \mbox{,}
    \label{eq:atmosphere scattering model}
\end{gathered}
\end{equation}
where $I$ and $J\in \mathbb{R}^{H \times W \times 3}$ denote the hazy and corresponding haze-free images of size $H \times W$. $A\in \mathbb{R}^{1}$ is the global atmospheric light, and the transmission map is given by $t=e^{-\beta d(x)}\in \mathbb{R}^{H \times W}$, where $\beta$ represents the haze density, $d(x)$ is the depth map, and $x$ is the pixel index.

However, as ASM is not sufficient to accurately simulate haze distribution in real-world scenarios, some studies~\cite{NHHaze20,NHHaze21,NHHaze23,IHAZE,OHAZE,DenseHaze} have focused on collecting paired real-world hazy and clean images to improve dehazing models' performance. Although dehazing models trained on real-world pairs perform better than those trained on synthetic pairs, they often suffer from significant performance drops when applied to unseen, real-world, hazy images from different domains. 

\begin{figure*}[t!]
    \begin{center}
    \includegraphics[width=1\textwidth]{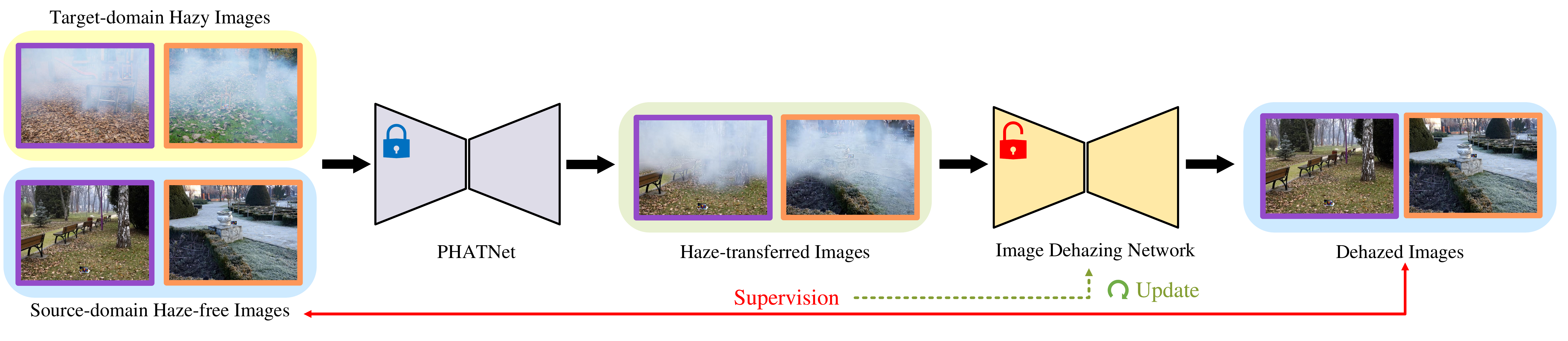}
    \end{center}
    \vspace{-0.2in}
    \caption{Illustration of the proposed Domain-Adaptive Dehazing framework. We introduce a Physics-guided Haze Transfer Network (PHATNet) which transfers haze patterns from hazy images in unseen target domains to clean images in the source domain. This process generates domain-adaptive fine-tuning sets, enabling dehazing models to be updated during testing for effective domain adaptation.} 
    \label{Figure_Introduction_2}
    \vspace{-0.1in}
\end{figure*} 

Since paired hazy and haze-free images for unseen target domains are typically unavailable during testing, some approaches~\cite{D4plus,HTFANet} utilize image-to-image translation techniques based on Generative Adversarial Networks (GANs)~\cite{NIPS2014_5ca3e9b1} to generate the paired images. However, GAN-based methods may encounter issues, such as mode collapse~\citep{asveegan17,MSGAN} and instability during optimization~\citep{NIPS2017_892c3b1c,8237566}. Moreover, they struggle to capture region-specific degradation patterns, such as non-uniform haze, commonly seen in real-world scenarios. Additionally, GANs lack effective domain adaptation capabilities, necessitating re-optimization of the generator to accommodate hazy images from unseen domains. These challenges motivate us to develop a flexible domain adaptation framework that allows dehazing models to effectively adjust to new target domains at test time.     

We present a novel domain adaptation framework to update dehazing models, enhancing their performance on unseen target-domain hazy images. Since haze-free counterparts of target-domain hazy images are typically unavailable during testing, it is essential to develop effective methods for generating paired hazy and clean images authentically reflecting target-domain haze patterns for test-time adaptation.
%
%
Our key observation is that retrieving haze patterns, especially thick haze, is generally easier than restoring haze-free content. This is because haze often forms a smooth, uniform, and semi-transparent layer that overlays the scene. ASM captures this phenomenon by simulating light scattering caused by atmospheric particles, resulting in a more predictable degradation in hazy images. Consequently, estimating haze is less challenging than reconstructing the original scene, which often contains intricate details obscured by the haze.
%
%
Based on this insight, we propose a \textbf{P}hysics-guided \textbf{HA}ze \textbf{T}ransfer \textbf{Net}work (PHATNet), designed to transfer haze patterns from unseen target domains to source-domain clean images. Through this process, PHATNet generates domain-adaptive fine-tuning sets with paired hazy/haze-free images, enabling effective test-time updates for dehazing models, as shown in Figure~\ref{Figure_Introduction_2}.

PHATNet leverages ASM guidance to transfer haze patterns extracted from target-domain hazy images. 
However, the standard ASM often fails to accurately represent real-world haze distributions due to its sensitivity to scene content, such as depth variations.
To address this, we propose a novel Parametric Haze Disentanglement and Transfer (PHDT) module that extends ASM to the latent space to disentangle haze-related and content-related features, facilitating their recombination to generate high-quality haze-transferred images. 
This latent space extension also enables PHATNet to mitigate ghosting artifacts caused by scene depth variations.
%
Additionally, we propose a Haze-Transfer-Consistency loss and a Content-Leakage Loss to further prevent PHATNet from leaking content-related features from hazy images.


PHATNet offers several advantages for domain adaptation in real-world dehazing tasks. First, by introducing ASM as an inductive bias for haze formation, PHATNet can accurately disentangle and transfer haze patterns from target-domain hazy images. This precision is achievable because haze exhibits greater regularity than natural scenes within the ASM parametric domain. Additionally, the ASM-guided approach ensures that the haze transfer and subsequent augmentation processes are physically interpretable. Second, PHATNet facilitates the augmentation of ample domain-specific fine-tuning data, effectively mitigating domain gaps and promoting domain-specific adaptation for dehazing models. Third, by utilizing an offline fine-tuning process, PHATNet ensures that dehazing models do not incur increased latency during testing after adaptation. 

Our key contributions are summarized as follows: 
\begin{itemize}
    \item We propose a flexible domain adaptation framework, called PHATNet, which disentangles and transfers haze patterns from target to source domains. This approach generates adaptive fine-tuning sets to update dehazing models, enhancing their performance during testing.
    \item Recognizing that extracting haze patterns is generally easier than restoring haze-free scene content from hazy images, we devise a Parametric Haze Disentanglement and Transfer (PHDT) module. This module, combined with a Haze-Transfer-Consistency Loss and a Content-Leakage Loss, enables effective haze pattern transfer.
    \item Extensive experimental results demonstrate our approach significantly boosts existing dehazing models across seven benchmark real-world image dehazing datasets. 
\end{itemize}

\section{Related Work}

\begin{figure*}[t!]
    \begin{center}
    \includegraphics[width=1\textwidth]{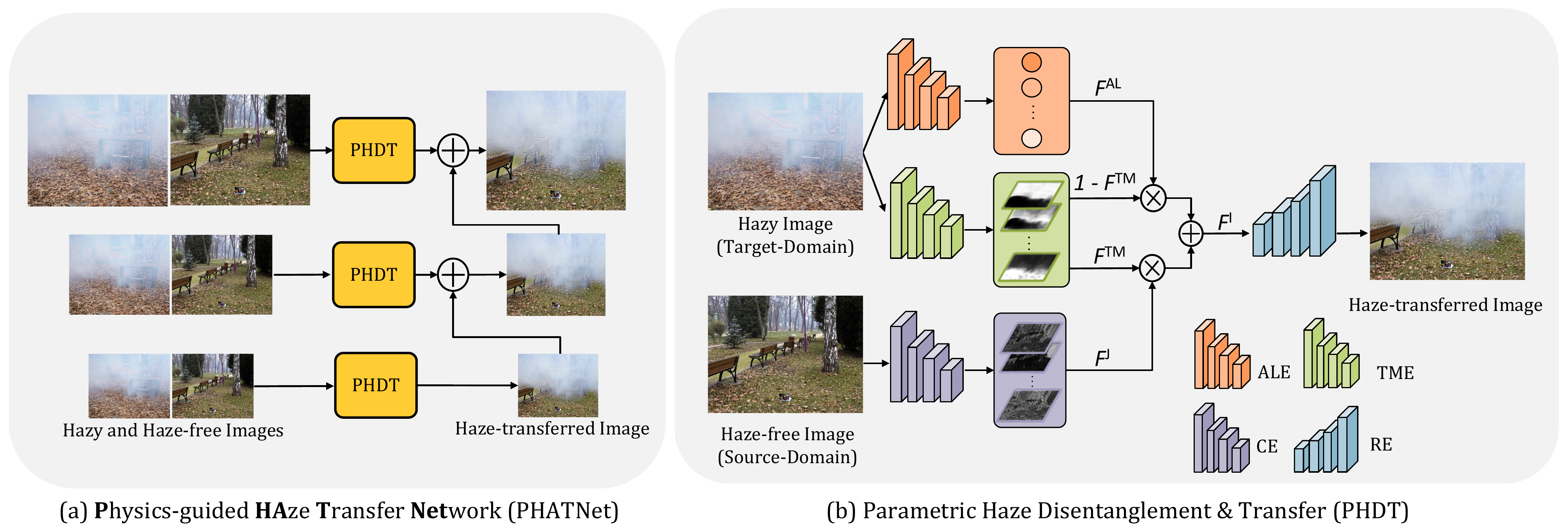}
    \end{center}
    \vspace{-0.2in}
    \caption{Architecture of the proposed Physics-guided Haze Transfer Network (PHATNet). (a) PHATNet is a multi-scale network composed of the proposed Parametric Haze Disentanglement and Transfer (PHDT). (b) PHDT is a dual-branch network where the top branch disentangles haze-related features--- specifically, atmospheric-light features $F^\mathrm{AL}$ by Atmospheric Light Encoder (ALE) and transmission-map features $F^\mathrm{TM}$ by Transmission Maps Encoder (TME)--- from a hazy image. In contrast, the bottom branch extracts content-related features $F^J$ by Content Encoder (CE) from a haze-free image. Guided by the Atmospheric Scattering Model (ASM), PHDT then combines these features in the ASM parametric domain to produce the final haze-transferred image by Rehazing Encoder (RE).}
    \label{fig:Architecture}
    \vspace{-0.05in}
\end{figure*} 

\subsection{Image Dehazing}
Image dehazing has made significant progress with the development of deep learning. Several studies~\cite{Ren_ECCV_2016,7539399,8237773,9157136,Hardgan} have aimed to enhance dehazing performance by leveraging prior knowledge from ASM. For instance, It was proposed in~\cite{Ren_ECCV_2016,7539399} to estimate the transmission map of a hazy image for dehazing. Deng~\etal~\cite{Hardgan} proposed a haze-aware representation distillation module by reformulating ASM. Other approaches~\cite{Frequency,9010659,8953692,Hong_2020_CVPR,MSBDN-DFF,qin2020ffa,Wu_2021_CVPR,Zheng_2021_CVPR,shen2023mitnet,10378631,song2023vision,cui2023focal} attempted to directly learn a hazy-to-clean mapping. For example, Guo~\etal~\cite{guo2022dehamer} proposed a transformer-based dehazing model with transmission-aware position embedding. Cui~\etal~\cite{cui2023focal} devised a dual-domain selection mechanism to amplify the response of important spatial and frequency regions. Shen~\etal~\cite{shen2023mitnet} presented a triplet interaction network to aggregate cross-domain, cross-scale, and cross-stage features. Although many architectural improvements have been made, these methods often fail to handle real-world hazy images across various domains, limiting their applicability in practical scenarios.

\subsection{Domain Adaptation for Image Restoration}
In addition to enhancing restoration performance through architectural design, several studies have focused on improving the performance of restoration models via data augmentation~\cite{D4plus,Wu_2024_IDBlau,HTFANet} and domain adaptation~\cite{9878808,chen2024promptbased,he2024domain,Liu_2022_BMVC} for image restoration.
In real-world hazy environments, haze patterns can be unpredictable, exhibiting a variety of shapes, color distortions, and contrast attenuation. This variability negatively impacts the performance of dehazing models when tested in different domains. To bridge the domain gap in image dehazing, several methods~\cite{9878808,HTFANet,D4plus,chen2024promptbased} have been proposed.
Since hazy-clean image pairs are unavailable during testing, Liu~\etal~\cite{9878808} proposed a helper network that reconstructs the hazy input from the dehazed output, allowing for the use of cycle-consistency loss to update dehazing models during testing. However, because the cycle reconstruction process functions as an auxiliary task, it limits the potential improvement of dehazing performance at inference.
Another approach to haze synthesis involves using GANs to generate hazy-clean image pairs that mimic haze distributions in target domains. For example, Li~\etal~\cite{HTFANet} and Yang~\etal~\cite{D4plus} applied ASM and GANs to transfer real-world haze patterns to clean images. However, GAN-based methods often suffer from mode collapse, 
limiting their ability to accurately transfer region-specific haze patterns. To overcome these limitations, Chen~\etal~\cite{chen2024promptbased} proposed to generate visual prompts that align with haze distributions in target domains through patch-based image-level normalization. However, these visual prompts may contain patch artifacts, reducing the fidelity of haze-transferred images.
%

\section{Proposed Method}

To address these challenges, we introduce the Physics-guided Haze Transfer Network (PHATNet), designed to transfer haze patterns from hazy images in unseen target domains to haze-free images in the source domain. This transfer creates domain-adaptive fine-tuning sets, enabling dehazing models to adapt effectively to new domains. At the core of PHATNet is the novel Parametric Haze Disentanglement and Transfer (PHDT) module, which disentangles haze patterns from a hazy image and scene content from a haze-free image and then combines them within the ASM parametric domain. To strengthen PHDT’s capacity to separate haze features from scene content, we introduce the Haze-Transfer-Consistency and Content-Leakage loss functions, enabling precise haze transfer. This section details the design of PHDT, PHATNet's architecture, the loss functions, and the domain adaptation process. 

\subsection{Parametric Haze Disentanglement \& Transfer}
To transfer haze patterns from a hazy image $I^H$ to a haze-free image $I^C$, where $I^H, I^C\in \mathbb{R}^{H \times W \times 3}$, we introduce PHDT with ASM guidance. As illustrated in Figure~\ref{fig:Architecture}, PHDT operates as a dual-branch network: the top branch extracts \textit{haze features} from the hazy image, while the bottom branch isolates \textit{content features} from the haze-free image. Using guidance from ASM, PHDT then fuses these features to produce the final haze-transferred image.    

In the top branch, we introduce the Atmospheric Light Encoder (ALE) and Transmission Maps Encoder (TME) to disentangle atmospheric-light features $F^\mathrm{AL}\in \mathbb{R}^{128}$ and transmission-map features $F^\mathrm{TM}\in \mathbb{R}^{\frac{H}{8} \times \frac{W}{8} \times 128}$ from the hazy image $I^H$ as follows:  
\begin{equation}
    F^\mathrm{AL} = \mathrm{exp}\{-{\mathrm{ALE}(I^H)}\}\mbox{,} 
\end{equation}
\begin{equation}
    F^\mathrm{TM} = \mathrm{exp}\{-{\mathrm{TME}(I^H)}\}\mbox{,}   
\end{equation}
where we apply the ReLU activation function at the end of the ALE and TME to clip negative values to zero, ensuring that the output features are non-negative. Following this, we use the function $e^{-x}$ to normalize $F^\mathrm{AL}$ and $F^\mathrm{TM}$ within the range $[0,1]$, maintaining consistency with the physical interpretation of ASM. 

\begin{figure}[t!]
    \begin{center}
    \includegraphics[width=1\columnwidth]{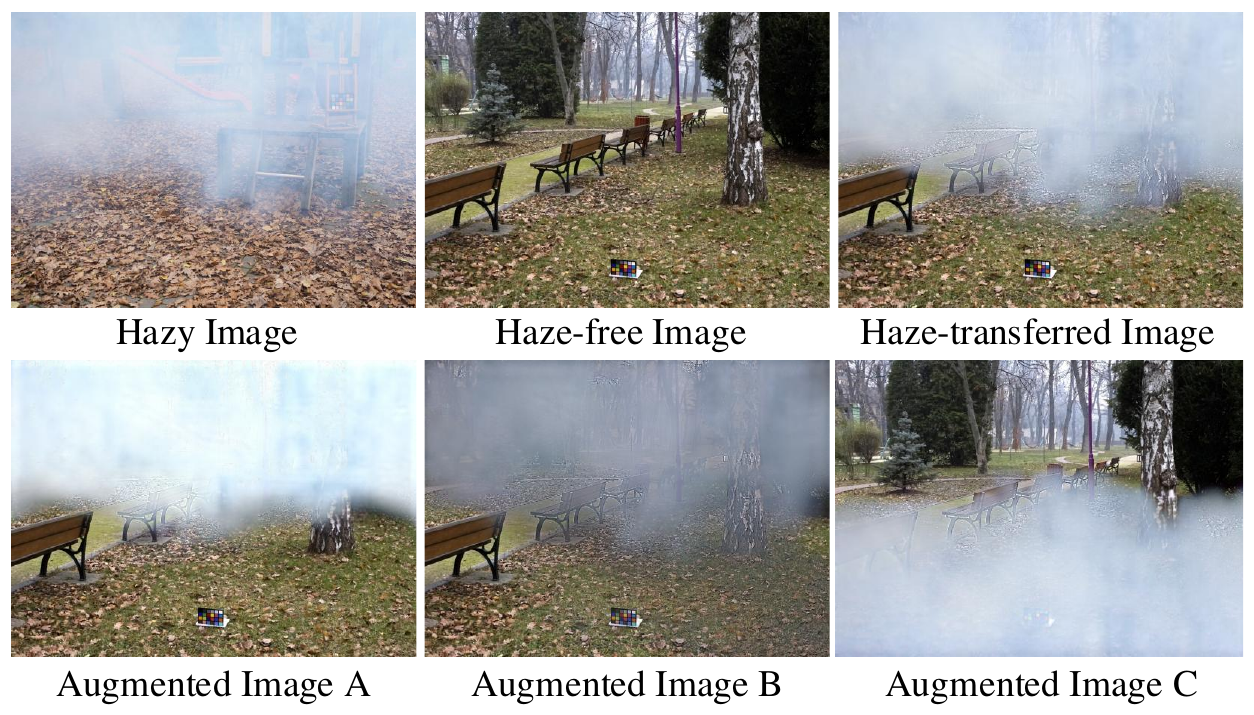}
    \end{center}
    \vspace{-0.2in}
    \caption{Illustration of PHDT-based hazy image augmentation. Beyond generating the haze-transferred image, we can apply gamma correction to  $F^\mathrm{TM}$ to adjust haze density, either increasing or decreasing it to further augment images $A$ and $B$. Additionally, we can vertically flip $F^\mathrm{TM}$ to augment image $C$.} 
    \label{Figure_Augmented_Image}
    \vspace{-0.1in}
\end{figure}

The bottom branch employs Content Encoder (CE) to disentangle content features $F^J\in \mathbb{R}^{\frac{H}{8} \times \frac{W}{8} \times 128}$ from $I^C$: 
\begin{equation}
    F^{J} = {\mathrm{CE}(I^C)}\mbox{.}   
\end{equation}

Next, we fuse $F^\mathrm{AL}$, $F^\mathrm{TM}$, and $F^J$ based on the ASM in Eq.~(\ref{eq:atmosphere scattering model}) to generate rehazed features $F^I\in \mathbb{R}^{\frac{H}{8} \times \frac{W}{8} \times 128}$ as 
\begin{equation}
    F^I = F^J \times F^\mathrm{TM} + F^\mathrm{AL} \times (1-F^\mathrm{TM})
    \mbox{.}    
    \label{eq:FLASM}
\end{equation}
Unlike the standard ASM in Eq.~(\ref{eq:atmosphere scattering model}) that operates in the image domain, we generate transmission-map features $F^\mathrm{TM}$ in the latent space, forming haze-aware attention maps invariant to scene depth. This enables us to disentangle haze-related features across channels without introducing ghosting artifacts. 
%
Finally, we employ a Rehazing Encoder (RE) to generate the final haze-transferred image $I^O\in \mathbb{R}^{H \times W \times 3}$. This image combines the haze patterns from $I^H$ with content from $I^C$.
The ALE, TME, CE, and RE modules are constructed from CNN-based residual blocks; additional architectural details are available in the supplementary material.
%
By generating hazy images aligned with ASM physical principles, we can further manipulate haze-transferred images in the ASM parametric domain to augment additional target-domain hazy images. For example, applying gamma correction to $F^\mathrm{TM}$ allows for haze density adjustments, while vertically flipping  $F^\mathrm{TM}$ creates additional variations, as illustrated in the augmented images $A$, $B$, and $C$ in Figure~\ref{Figure_Augmented_Image}.

\subsection{Multi-scale Haze Transfer}
Given a hazy image $I^H$ from the target domain and a haze-free image $I^C$ from the source domain, PHATNet can generate a haze-transferred image $I^O$, which incorporates target-domain haze patterns from $I^H$ along with content information from $I^C$ as follows: $I^O=\mathrm{PHATNet}(I^{H},I^{C})$. Since haze patterns are often non-homogeneous and can vary in scales, as shown in Figure~\ref{fig:Architecture}(a), we employ a multi-scale PHDT structure in PHATNet to facilitate the transfer of haze patterns in the following coarse-to-fine manner:
\begin{equation}                                
    I^{O} =\mathrm{PHATNet}(I^{H},I^{C}) \\
    = \mathrm{PHDT}(I^{H}, I^{C}) + \mathrm{UP}(\mathrm{I^{O\downarrow}})\mbox{,} 
\end{equation}
where
\begin{equation}
    I^{O\downarrow} = \mathrm{PHDT}(I^{H\downarrow}, I^{C\downarrow}) + \mathrm{UP}(\mathrm{I^{O\downarrow\downarrow}})\mbox{,} 
\end{equation}
and
\begin{equation}
    I^{O\downarrow\downarrow} = \mathrm{PHDT}(I^{H\downarrow\downarrow}, I^{C\downarrow\downarrow})\mbox{,} 
\end{equation}
where $(\cdot)^{\downarrow}$ and $(\cdot)^{\downarrow\downarrow}$ denote the downsampling-by-two and downsampling-by-four operations, respectively, and $\mathrm{UP}(\cdot)$ represents the upsampling-by-two operation. 

\begin{figure}[t!]
    \begin{center}
    \includegraphics[width=1\columnwidth]{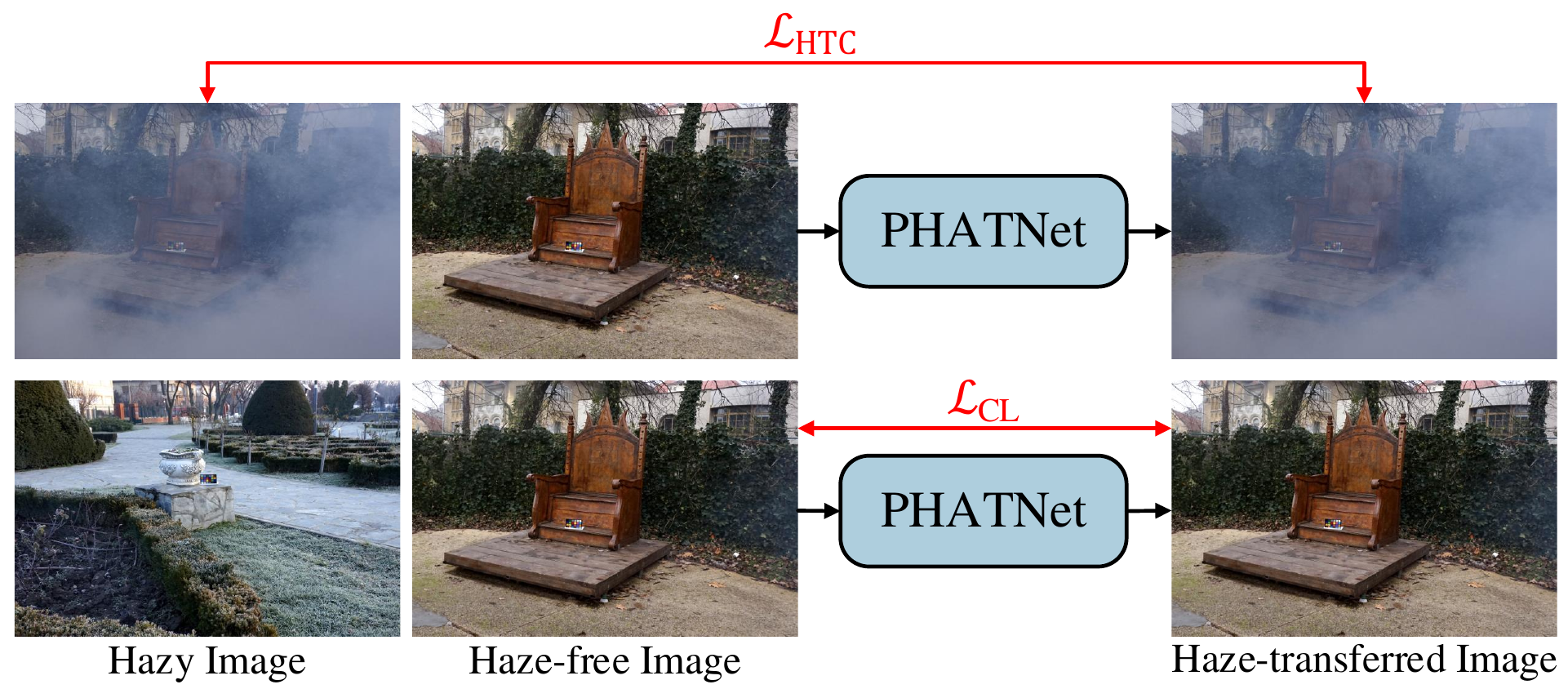}
    \end{center}
    \vspace{-0.2in}
    \caption{{Illustration of the loss functions for PHATNet. The top figure depicts the cyclic haze-transfer process of PHATNet, supervised by the Haze-Transfer-Consistency Loss ($\mathcal{L}_\mathrm{HTC}$). In the bottom figure, since the hazy image lacks haze content, the Content-Leakage Loss ($\mathcal{L}_\mathrm{CL}$) is applied to ensure that PHATNet generates a haze-transferred image identical to the original haze-free image.}}
    \label{Figure_CRL}
    \vspace{-0.1in}
\end{figure}

\subsection{Loss Functions}
We use paired real-world hazy and haze-free images $\{I^{H}_i, I^{C}_i\}^{N}_{i=1}$ in the source domain, consisting of $N$ pairs, to train PHATNet. The optimization process is driven by two key loss functions: the Haze-Transfer-Consistency Loss ($\mathcal{L}_\mathrm{HTC}$) and the Content-Leakage Loss ($\mathcal{L}_\mathrm{CL}$), as illustrated in Figure~\ref{Figure_CRL}. The Haze-Transfer-Consistency Loss $\mathcal{L}_\mathrm{HTC}$ assesses the difference between the haze-transferred image $I^{O}_i=\mathrm{PHATNet}(I^{H}_i,I^{C}_i)$ and its corresponding hazy reference, $I^{H}_i$. To generate the haze-transferred image $I^{O}_i$, PHATNet performs a cyclic haze-transfer process: haze features from $I^{H}_i$ are extracted in the ASM parametric domain, and then combined with scene content from haze-free image $I^{C}_i$, yielding $I^{O}_i$ in the image domain, ideally capturing the same haze patterns as $I^{H}_i$.   $ \mathcal{L}_\mathrm{HTC}$ is defined as
\begin{equation}
    \mathcal{L}_\mathrm{HTC}=\sum_{s=1}^{3} ||I^{O}_{i, s}-I^{H}_{i, s}||_1, 
\end{equation}
where $s$ denotes the scale index in the multi-scale network.

To isolate haze patterns exclusively from hazy images for haze transfer, it is essential that these haze patterns, extracted from hazy images $I^{H}_i$, remain disentangled from the underlying scene content. To achieve this, we introduce the Content-Leakage Loss, which minimizes content leakage from hazy images $I^{H}_i$ into the corresponding haze-transferred images. This method reconsiders the use of unpaired image translation without relying on GANs, which are often unstable in training. To prevent the atmospheric light and transmission map encoders from capturing content-related features, we use unpaired haze-free images as input to PHATNet, generating $I^O_{i,j} = \mathrm{PHATNet}(I^{C}_i, I^{C}_j)$, where $j \neq i$. In this setup, $I^O_{i,j}$ incorporates haze patterns from $I^{C}_i$ while preserving the scene content from $I^{C}_j$. Since $I^{C}_i$ is haze-free, $I^O_{i,j}$ should ideally only reflect the content of $I^{C}_j$. Thus, the Content-Leakage Loss can be defined as     
\begin{equation}
    \mathcal{L}_\mathrm{CL}= \sum_{s=1}^{3}||I^O_{i,j,s}-I^{C}_{j,s}||_1,
\end{equation}
%
Consequently, we optimize PHATNet using the total loss: 
\begin{equation}
    \mathcal{L}_\mathrm{total}= \mathcal{L}_\mathrm{HTC} + \mathcal{L}_\mathrm{CL}.
\end{equation}

\subsection{Domain Adaptation via Haze Transfer}
    After training PHATNet, we use it to transfer haze patterns from hazy images $\{\Tilde{I}^{H}_i\}^M_{i=1}$ in the target domain to haze-free images $\{I^{C}_j\}^N_{j=1}$ in the source domain, producing a domain-specific fine-tuning set $S=\{\Tilde{I}^{O}_{i,j},  I^{C}_j\}_{\forall i,j}$ with a total of $M \times N$ training pairs. Here, $\Tilde{I}^{O}_{i,j}=\mathrm{PHATNet}(\Tilde{I}^{H}_i,I^{C}_j)$ represents the haze-transferred image that combines haze patterns from $\Tilde{I}^{H}_i$ with the scene content from $I^{C}_j$, while $I^{C}_j$ serves as its corresponding haze-free version. We then fine-tune dehazing models on the generated domain-specific fine-tuning set $S$, enabling effective domain adaptation.

\section{Experiments}
\subsection{Datasets}
We utilize seven real-world image dehazing datasets in our experiments: NH-Haze20~\cite{NHHaze20}, NH-Haze21~\cite{NHHaze21}, HD-NH-Haze~\cite{NHHaze23}, DenseHaze~\cite{DenseHaze}, I-Haze~\cite{IHAZE}, O-Haze~\cite{OHAZE}, and RTTS~\cite{Resides}. Specifically, NH-Haze20, NH-Haze21, HD-NH-Haze, DenseHaze, I-Haze, and O-Haze provide paired hazy and haze-free images, with $55$, $25$, $40$, $55$, $30$, and $40$ pairs, respectively. In contrast, the RTTS dataset only contains $4,322$ hazy images without their haze-free versions. 
We conduct two experimental settings to assess the effectiveness of the proposed method, as summarized in Table~\ref{setting}. In \textbf{Setting1}, we use NH-Haze20 as the source domain for optimizing both PHATNet and the dehazing models. Consistent with prior studies~\cite{cui2023focal, guo2022dehamer, shen2023mitnet}, we employ the first $50$ image pairs for training and the last $5$ pairs for validation. The remaining datasets are used as target domains for performance evaluation. In \textbf{Setting2}, we use HD-NH-Haze as the source domain, training on the initial $35$ pairs and validating on the last $5$ pairs. The other datasets then serve as target domains for testing.

\begin{table}[t!]
\small
\centering
\setlength{\tabcolsep}{1mm}
\caption{Experimental settings applied in our experiments.}
\vspace{-0.1in}
\begin{tabular}{c|c|cc}
\noalign{\hrule height 1.0pt}
Setting        & Source Domain  &  \multicolumn{2}{c}{Target Domains}  \\
\noalign{\hrule height 1.0pt}
\multirow{3}{*}{\bf{Setting1}} & \multirow{3}{*}{NH-Haze20~\cite{NHHaze20}} & NH-Haze21~\cite{NHHaze21} & HD-NH-Haze~\cite{NHHaze23} \\ & & I-Haze~\cite{IHAZE} & O-Haze~\cite{OHAZE}  \\ & & DenseHaze~\cite{DenseHaze} & RTTS~\cite{Resides} \\
\noalign{\hrule height 1.0pt}
\multirow{3}{*}{\bf{Setting2}} & \multirow{3}{*}{HD-NH-Haze~\cite{NHHaze23}} & NH-Haze20~\cite{NHHaze20} & NH-Haze21~\cite{NHHaze21} \\ & & I-Haze~\cite{IHAZE} & O-Haze~\cite{OHAZE}  \\ & & DenseHaze~\cite{DenseHaze} & RTTS~\cite{Resides} \\
\noalign{\hrule height 1.0pt}
\end{tabular}
\label{setting}
\vspace{-0.2in}
\end{table}

\begin{table*}[t!]
\centering
\setlength{\tabcolsep}{2mm}
\caption{Evaluation results of dehazing performance under \textbf{setting1}. We use NH-Haze20~\cite{NHHaze20} as the source domain and evaluate the performance on target domains, including NH-Haze21~\cite{NHHaze21}, HD-NH-Haze~\cite{NHHaze23}, DenseHaze~\cite{DenseHaze}, I-Haze~\cite{IHAZE}, O-Haze~\cite{OHAZE}, respectively. ``Baseline'' and ``+PHATNet'' denote the dehazing performance without and with PHATNet.}
\vspace{-0.1in}
\footnotesize
\begin{tabular}{cc |cc| cc| cc| cc| cc | cc}
\hline\hline
& & \multicolumn{2}{c|}{\bf{NH-Haze21}} & \multicolumn{2}{c|}{\bf{HD-NH-Haze}} & \multicolumn{2}{c|}{\bf{DenseHaze}} & \multicolumn{2}{c|}{\bf{I-Haze}} & \multicolumn{2}{c|}{\bf{O-Haze}} & \multicolumn{2}{c}{\bf{Average}} \\
\multicolumn{2}{c|}{Model}  & PSNR   & SSIM   & PSNR  & SSIM & PSNR  & SSIM & PSNR  & SSIM & PSNR  & SSIM & PSNR  & SSIM   \\ \hline
\multirow{2}{*}{FocalNet~\cite{cui2023focal}} & Baseline &   16.45    &  0.635   &  14.76  & 0.509  &   14.80    &  \bf0.434   & 16.13  & 0.588  & 19.10  & 0.706 & 16.25 & 0.574 \\ 
& +PHATNet &   \bf 16.90    &  \bf 0.654   &  \bf15.87  &   \bf0.570    &  \bf15.60  &   0.423 &  \bf16.96  &   \bf0.602 & \bf20.01 & \bf 0.711 & \bf17.07 & \bf0.592 \\ \hline

\multirow{2}{*}{Dehamer~\cite{guo2022dehamer}} & Baseline &  16.46    &  0.619   &  14.30   &   0.457 &   14.56    &  0.436   &   17.05  & 0.573 &   19.28  & 0.696 & 16.33 & 0.556 \\ & +PHATNet &   \bf17.26   &  \bf0.636   &  \bf15.22  &  \bf0.537 &   \bf15.42   &  \bf0.438   &  \bf 17.85 &   \bf0.625 & \bf20.70 & \bf0.697 & \bf 17.29  &  \bf 0.587 \\ \hline

\multirow{2}{*}{MITNet~\cite{shen2023mitnet}} & Baseline &   14.21    &  0.516   &  11.60   &   0.355 &   13.67    &  0.420   &  15.72   &   0.540 &  18.96   &  0.701 & 14.83 & 0.506 \\ & +PHATNet &   \bf16.21   &  \bf0.650  &  \bf14.04  &   \bf0.466 &   \bf14.73   &  \bf0.428   &  \bf 16.54  &   \bf0.557 & \bf20.02 & \bf0.708 & \bf16.31 & \bf0.562  \\ \hline

\multirow{2}{*}{SGDN~\cite{fang2025guided}} & Baseline &   14.79    &  0.560   &  12.40   &   0.424 &   13.71    &  0.438   &  14.74   &   0.568 &  18.55   &  0.701 & 14.84 & 0.538 \\ & +PHATNet &   \bf17.64   &  \bf0.677  &  \bf16.67  &   \bf0.609 &   \bf14.21   &  \bf0.438   &  \bf 16.03  &   \bf0.599 & \bf19.31 & \bf0.705 & \bf 16.77 & \bf 0.606 \\ \hline

\hline\hline
\end{tabular}
\label{Tab:seeting1}
\end{table*}

\begin{table*}[t!]
\centering
\setlength{\tabcolsep}{2mm}
\caption{Evaluation results of dehazing performance on \textbf{setting2}. We use HD-NH-Haze~\cite{NHHaze23} as the source domain and evaluate on target domains, including NH-Haze20~\cite{NHHaze20}, NH-Haze21~\cite{NHHaze21}, DenseHaze~\cite{DenseHaze}, I-Haze~\cite{IHAZE}, O-Haze~\cite{OHAZE}, respectively. ``Baseline'' and ``+PHATNet'' denote the dehazing performance without and with PHATNet.}
\vspace{-0.1in}
\footnotesize
\begin{tabular}{cc |cc| cc| cc| cc| cc | cc}
\hline\hline
& & \multicolumn{2}{c|}{\bf{NH-Haze20}} & \multicolumn{2}{c|}{\bf{NH-Haze21}} & \multicolumn{2}{c|}{\bf{DenseHaze}} & \multicolumn{2}{c|}{\bf{I-Haze}} & \multicolumn{2}{c|}{\bf{O-Haze}} & \multicolumn{2}{c}{\bf{Average}} \\
\multicolumn{2}{c|}{Model}  & PSNR   & SSIM   & PSNR  & SSIM & PSNR  & SSIM & PSNR  & SSIM & PSNR  & SSIM & PSNR  & SSIM  \\ \hline
\multirow{2}{*}{FocalNet~\cite{cui2023focal}} & Baseline &   15.12    &  0.468   &  19.90  & 0.799  &   11.34    &  0.323   & 13.92  & 0.565  & 16.64  & 0.558 & 15.38 & 0.543 \\ 
& +PHATNet &   \bf 15.77    &  \bf 0.505   &  \bf20.06  &   \bf0.808  &  \bf12.95  &  \bf0.323 &  \bf14.64  &   \bf0.593 & \bf17.66 & \bf 0.586 & \bf16.22 & \bf0.563 \\ \hline

\multirow{2}{*}{Dehamer~\cite{guo2022dehamer}} & Baseline &  14.82    &  0.475   &  18.62   &   0.748 &   12.57    &  0.320   &   16.06  & 0.569 &   17.67  & 0.576 & 15.95 & 0.538 \\ 
& +PHATNet &   \bf15.10   &  \bf0.478   &  \bf18.73  &   \bf0.747 &   \bf13.61   &  \bf0.331   &  \bf 16.65 &   \bf0.570 & \bf18.95 & \bf0.585 
& \bf 16.61 & \bf 0.542 \\ \hline

\multirow{2}{*}{MITNet~\cite{shen2023mitnet}} & Baseline &   11.61    &  0.385   &  19.30  &   0.775 &   8.6    &  0.285   &  9.99   &  0.548 &  12.70   &   0.502 & 12.44 & 0.499 \\ 
& +PHATNet &   \bf14.02   &  \bf0.477  &  \bf19.64  &   \bf0.783 &   \bf11.22   &  \bf0.292   &  \bf 15.14  &   \bf0.550 & \bf18.94 & \bf0.561 & \bf 15.79 & \bf0.533 \\ \hline

\multirow{2}{*}{SGDN~\cite{fang2025guided}} & Baseline &   11.87    &  0.414   &  19.95   &   0.810 &   8.29    &  0.232   &  11.01   &   0.524 &  13.84   &  0.522 & 12.99 & 0.500 \\ & +PHATNet &   \bf14.71   &  \bf0.502  &  \bf20.33  &   \bf0.815 &   \bf11.92   &  \bf0.310   &  \bf 15.57  &   \bf0.572 & \bf18.34 & \bf0.572 & \bf 16.17 & \bf 0.554 \\ \hline

\hline\hline
\end{tabular}
\label{Tab:seeting2}
\end{table*}

\begin{table*}[t!]
\centering
\setlength{\tabcolsep}{1mm}
\caption{Evaluation results on the RTTS~\cite{Resides} dataset, where we respectively use NH-Haze20~\cite{NHHaze20} and HD-NH-Haze~\cite{NHHaze23} as source domains.}
\vspace{-0.1in}
\small
\begin{tabular}{cc|cc|cc|cc|cc}
\hline\hline
&  & \multicolumn{2}{c|}{FocalNet~\cite{cui2023focal}} & \multicolumn{2}{c|}{Dehamer~\cite{guo2022dehamer}} & \multicolumn{2}{c|}{MITNet~\cite{shen2023mitnet}} & \multicolumn{2}{c}{SGDN~\cite{fang2025guided}} \\ 
& & Baseline & +PHATNet & Baseline & +PHATNet & Baseline & +PHATNet & Baseline & +PHATNet \\
\noalign{\hrule height 1.0pt}

\multirow{2}{*}{NH-Haze20 $\rightarrow$ RTTS} & NIQE ($\downarrow$) & 4.19 & \bf3.92 & 4.20 & \bf3.82 & 4.42 & \bf4.33 & 4.30 & \bf 3.85 \\ & BRISQUE ($\downarrow$) & 34.88 & \bf28.34 & 31.94 & \bf 24.15 & 33.67 & \bf 31.99 & 34.27 & \bf 22.21 \\
\noalign{\hrule height 1.0pt}

\multirow{2}{*}{HD-NH-Haze $\rightarrow$ RTTS} & NIQE ($\downarrow$) & 4.45 & \bf4.40 & 4.49 & \bf4.33 & 4.69 & \bf4.62 & 4.74 & \bf 4.58 \\ & BRISQUE ($\downarrow$) & 32.78 & \bf28.93 & 34.67 & \bf 23.74 & 30.21 & \bf 28.84 & 29.71 & \bf 27.63  \\
\hline\hline
\end{tabular}
\label{Tab:RTTS}
\vspace{-0.1in}
\end{table*}

\subsection{Implementation Details}
\paragraph{PHATNet} PHATNet is optimized on source-domain training pairs over 1,000 epochs, with a batch size of 1 and an image resolution of $1600\times1200$ on an Nvidia A5000 GPU. We employ the Adam optimizer with an initial learning rate of $10^{-4}$, which decays to $10^{-7}$ following a cosine annealing schedule. 

\vspace{-0.1in}
\paragraph{Dehazing Models} We evaluate the effectiveness of PHATNet using four SOTA dehazing models: FocalNet~\cite{cui2023focal}, Dehamer~\cite{guo2022dehamer}, MITNet~\cite{shen2023mitnet}, and the latest SGDN~\cite{fang2025guided}. For each model, we utilize its source-domain pre-trained weights, if available; if not, we retrain the model on the source domain's training set using its default settings. During the testing phase, each dehazing model is fine-tuned for one epoch on the fine-tuning set generated for each target domain.   

\subsection{Experimental Results}
\paragraph{Quantitative Results} We compare the dehazing performance of four SOTA baseline models and their PHATNet-enhanced versions in Tables~\ref{Tab:seeting1} and~\ref{Tab:seeting2}, where ``Baseline'' indicates performance without PHATNet-augmented data, while ``+PHATNet'' means performance with it.
Table~\ref{Tab:seeting1} compares the PSNR performance under \textbf{Setting1}, showing substantial gains across all models using PHATNet. Specifically, the average PSNR gains enhanced by PHATNet across the five datasets are $0.82$ dB for FocalNet, $0.96$ dB for Dehamer, $1.48$ dB for MITNet, and $1.93$ dB for SGDN.

%
Similarly, Table~\ref{Tab:seeting2} demonstrates that under \textbf{Setting2}, PHATNet consistently enhances the average PSNR performance across all datasets, with gains of $0.84$ dB for FocalNet, $0.66$ dB for Dehamer, $3.35$ dB for MITNet, and $3.18$ dB for SGDN. 
%
Table~\ref{Tab:RTTS} further demonstrates the effectiveness of PHATNet on RTTS, a large real-world dataset without haze-free reference images. Employing two commonly-used no-reference image quality metrics, NIQE~\cite{NIQE} and BRISQUE~\cite{BRISQUE}, we show that PHATNet consistently enhances performance on RTTS.     
\vspace{-0.2in}

\begin{figure*}[t!]
    \begin{center}
        \includegraphics[width=1\textwidth]{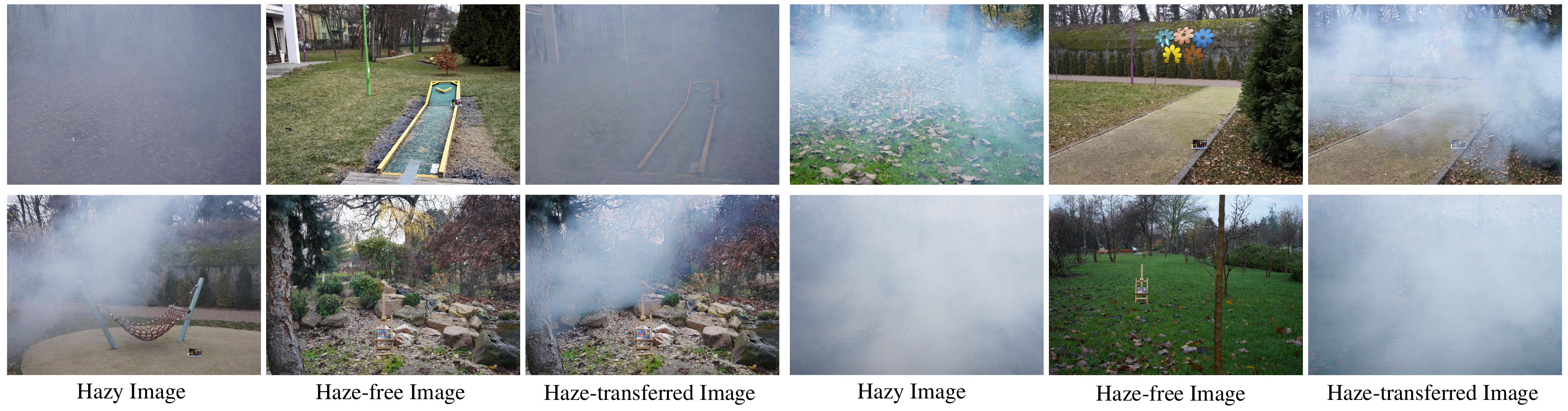}
    \end{center}
    \vspace{-0.2in}
    \caption{Qualitative results of haze-transferred images. In the top row, we transfer haze patterns from O-Haze~\cite{OHAZE} (top-left figure) and HD-NH-Haze~\cite{NHHaze23} (top-right figure) to haze-free images from NH-Haze20~\cite{NHHaze20}. In the bottom row, we transfer haze patterns from NH-Haze20~\cite{NHHaze20} (bottom-left figure) and DenseHaze~\cite{DenseHaze} (bottom-right figure) to haze-free images from HD-NH-Haze~\cite{NHHaze23}.} 
    \label{Rehaze}
    \vspace{-0.1in}
\end{figure*}

\begin{figure*}[t!]
    \begin{center}
        \includegraphics[width=1\textwidth]{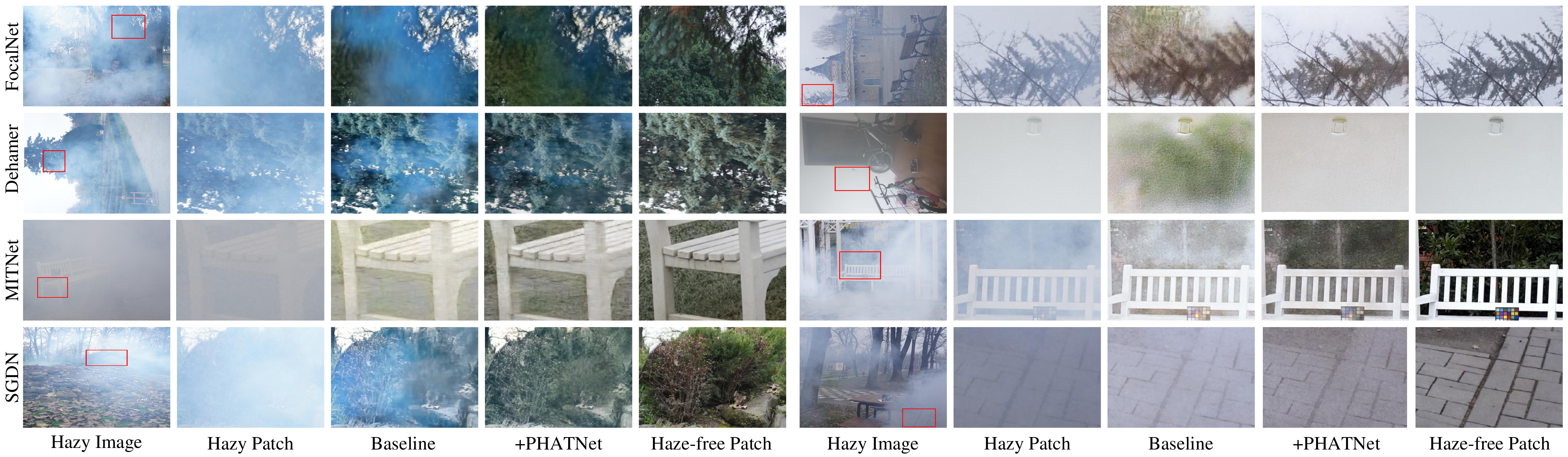}
    \end{center}
    \vspace{-0.2in}
    \caption{Qualitative performance comparisons of dehazed images conducted under \textbf{Setting1} and \textbf{Setting2}. Left: We demonstrate the dehazed images on HD-NH-Haze~\cite{NHHaze23}, NH-Haze21~\cite{NHHaze21}, and DenseHaze~\cite{DenseHaze} datasets under \textbf{Setting1}. Right: We demonstrate the dehazed images on O-Haze~\cite{OHAZE}, I-Haze~\cite{IHAZE}, and NH-Haze20~\cite{NHHaze20} datasets under \textbf{Setting2}.} 
    \label{Dehazed_Vis}
    \vspace{-0.1in}
\end{figure*}

\paragraph{Qualitative Results}
Figure~\ref{Rehaze} displays haze-transferred images generated by PHATNet, with haze-free images from NH-Haze20 and HD-NH-Haze training sets in the top and bottom rows, respectively. PHATNet effectively transfers region-specific haze patterns from unseen target domains to source-domain haze-free images, capturing realistic haze characteristics. Figure~\ref{Dehazed_Vis} presents dehazed images produced with and without using PHATNet-augmented data under \textbf{Setting1} and \textbf{Setting2}, respectively. These figures showcase PHATNet's ability to enhance visual quality by reducing hazy artifacts, highlighting PHATNet's robust domain adaptation capabilities. Additional qualitative results can be found in the supplementary material.

\subsection{Ablation Studies}
We evaluate the impact of PHATNet-augmented data on dehazing performance after adaptation, specifically using FocalNet as the dehazing model under \textbf{Setting1}. First, we examine the contributions of the ALM and TME in the multi-scale PHDT framework within PHATNet. Next, we assess how the Content-Leakage Loss affects PHATNet's performance. At last, we compare PHATNet with existing domain adaptation approaches~\cite{9878808,chen2024promptbased} and haze synthesis methods~\cite{HTFANet,D4plus} for dehazing. In the following tables, ``Baseline'' refers to FocalNet trained solely on the source-domain training set without PHATNet-augmented data.     

\begin{table}[t!]
\small
\centering
\setlength{\tabcolsep}{4mm}
\caption{PSNR (dB) with ALE and TME within PHDT.}
\vspace{-0.1in}
\begin{tabular}{c|ccc}
\noalign{\hrule height 1.0pt}
& ALE & TME & PSNR  \\
\noalign{\hrule height 1.0pt}
Baseline &  &  & 16.25 \\
+CNN (Concate) &  &  & 16.62 \\
+ALE & $\checkmark$ &  & 16.46 \\
+TME &  & $\checkmark$   & 16.80  \\
Ours & $\checkmark$ & $\checkmark$ & \bf 17.07 \\
\noalign{\hrule height 1.0pt}
\end{tabular}
\vspace{-0.1in}
\label{FASM_components}
\end{table}

\vspace{-0.25in}
\paragraph{Component analysis of PHATNet} 
Table~\ref{FASM_components} evaluates the individual and combined contributions of the Atmospheric Light Encoder (``+ALE'') and Transmission Map Encoder (``+TME''). 
``+CNN (Concate)'' denotes that features extracted from both hazy and haze-free images are concatenated for fusion. Experimental results demonstrate that integrating ALE and TME performs better than using either component alone or the feature concatenation strategy.
Table~\ref{mult_scale} indicates that employing a 3-stage PHDT structure in PHATNet optimizes dehazing performance, achieving the highest PSNR.  

\begin{figure}[t!]
    \begin{center}
    \includegraphics[width=0.8\columnwidth]{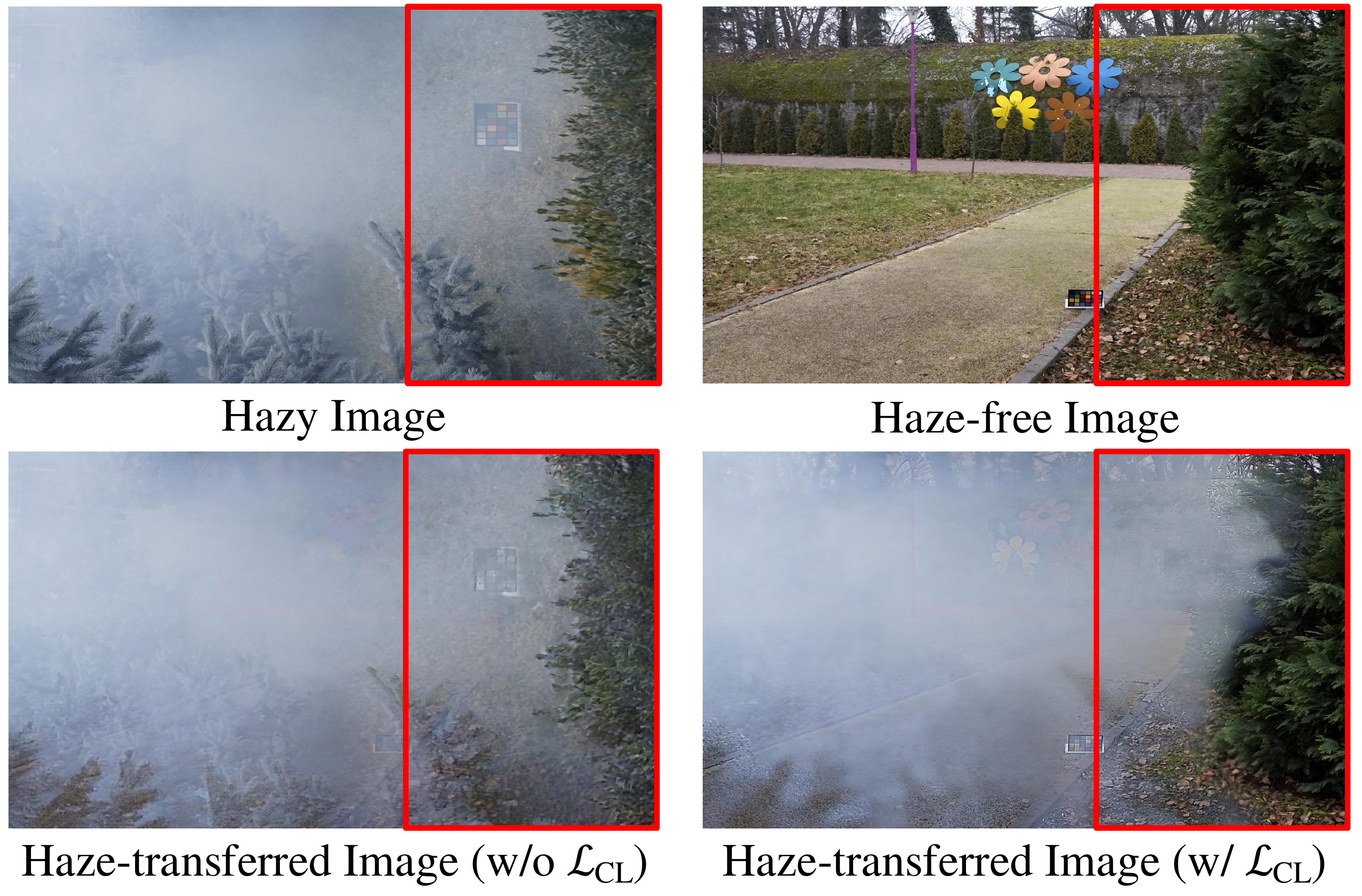}
    \end{center}
    \vspace{-0.2in}
    \caption{Illustration of haze-transferred images generated without and with the content-leakage loss, denoted ``w/o $\mathcal{L}_\mathrm{CL}$'' and ''w/ $\mathcal{L}_\mathrm{CL}$'', within PHATNet, where we use the hazy image from O-Haze~\cite{OHAZE} and haze-free image from NH-Haze20~\cite{NHHaze20}.} 
    \label{CRL_Vis}
    \vspace{-0.2in}
\end{figure}

\vspace{-0.1in}
\paragraph{Effectiveness of content-leakage loss} 
Table~\ref{CRL} presents the effectiveness of Content-Leakage (CL) Loss, $\mathcal{L}_\mathrm{CL}$, comparing the performance of PHATNet without CL Loss (``w/o $\mathcal{L}_\mathrm{CL}$'') and with CL Loss (``w/ $\mathcal{L}_\mathrm{CL}$''). The $\mathcal{L}_\mathrm{CL}$ mitigates content leakage from hazy images into ALE and TME, generating more accurate haze-transferred images and enhancing dehazing performance. While PHATNet without $\mathcal{L}_\mathrm{CL}$ improves dehazing results on NH-Haze21, HD-NH-Haze, DenseHaze, and I-Haze datasets, incorporating $\mathcal{L}_\mathrm{CL}$ yields further PSNR gains across all datasets. This effect is particularly notable on O-Haze, which contains sparse haze, making the model more susceptible to content interference during feature disentanglement. Figure~\ref{CRL_Vis} provides visual comparisons of haze-transferred images generated with and without $\mathcal{L}_\mathrm{CL}$.    

\begin{table}[t!]
\small
\centering
\setlength{\tabcolsep}{2mm}
\caption{Impact of the stage number of PHDT on performance.}
\vspace{-0.1in}
\begin{tabular}{c|cccc}
\noalign{\hrule height 1.0pt}
  & Baseline & 1-stage & 2-stage & 3-stage  \\
\noalign{\hrule height 1.0pt}
PSNR & 16.25 & 16.68  & 16.95  &  \bf 17.07 \\ 
\noalign{\hrule height 1.0pt}
\end{tabular}
\vspace{-0.2in}
\label{mult_scale}
\end{table}

\begin{table}[t!]
\small
\centering
\setlength{\tabcolsep}{4mm}
\caption{Impact of Content-Leakage Loss ($\mathcal{L}_\mathrm{CL}$) on PHATNet's PSNR performance (dB).}
\vspace{-0.1in}
\begin{tabular}{c|ccc}
\noalign{\hrule height 1.0pt}
& Baseline & w/o $\mathcal{L}_\mathrm{CL}$ & w/ $\mathcal{L}_\mathrm{CL}$ \\ 
\noalign{\hrule height 1.0pt}
NH-Haze21 & 16.45 & \underline{16.68} & \bf16.90   \\
HD-NH-Haze &  14.76 & \underline{15.48} & \bf15.87 \\
DenseHaze &  14.80 & \underline{15.14} & \bf15.60  \\
I-Haze  & 16.13 & \underline{16.37} & \bf16.96  \\
O-Haze & \underline{19.10} & 18.04 & \bf20.01  \\
\noalign{\hrule height 1.0pt}
\end{tabular}
\label{CRL}
\end{table}

\vspace{-0.1in}
{\paragraph{Comparison with competing methods} 
Table~\ref{Comparison} compares the performance and complexity of PHATNet against two domain-adaptation methods, TMD~\cite{9878808} and PTTD~\cite{chen2024promptbased}, and two haze-synthesis methods, HTFANet~\cite{HTFANet} and D4+~\cite{D4plus}. For a fair comparison, we use FocalNet as the dahzing model and follow \textbf{Setting1}, using NH-Haze20 training set to optimize the competing methods and reporting the average PSNR. We also compare the average runtime needed for augmenting a hazy image for the adaptation process. For the GAN-based HTFANet and D4+, both paired and unpaired data in the training set are used for training. 
Table~\ref{Comparison} shows that PHATNet outperforms all competing methods in PSNR, demonstrating its effectiveness in domain adaptation for real-world image dehazing. With 26 million parameters, PHATNet requires 0.153 seconds to generate a $1600\times1200$ haze-transferred image on an Nvidia A5000 GPU, which is efficient and effective compared to the others.
In addition, since the adaptation process is performed offline, there is no increase in latency for dehazing models during inference.

The qualitative comparison in Figure~\ref{Haze_Transfer_Comparison_Crop} further demonstrates PHATNet's strong ability to generate realistic haze-transferred images. In particular, since HTFANet utilizes vanilla ASM to transfer haze patterns, it inevitably leaks content-related features in the transmission map, as shown in Figure~\ref{Transmission_crop}. In contrast, PHATNet successfully disentangles haze-related features by extending ASM to the latent space, effectively mitigating ghosting artifacts caused by scene depth variations.



\vspace{-0.1in}
\paragraph{Discussions} 
Beyond haze transfer, PHATNet can further augment hazy images through gamma correction and flipping in the ASM parametric domain (see Figure~\ref{Figure_Augmented_Image}). However, achieving comprehensive hazy image augmentation in the parametric domain remains challenging due to limited prior knowledge of real-world haze distributions. This limitation highlights a valuable avenue for future research.

\begin{table}[t!]
\small
\centering
\setlength{\tabcolsep}{1mm}
\caption{Comparison of performance and complexity between our method, domain-adaptation methods, TMD~\cite{9878808} and PTTD~\cite{chen2024promptbased}, and haze-synthesis methods, HTFANet~\cite{HTFANet} and D4+~\cite{D4plus}.}
\vspace{-0.1in}
\begin{tabular}{c|cccccc}
\noalign{\hrule height 1.0pt}
& Baseline & TMD & HTFANet & D4+ & PTTD & Ours  \\
\noalign{\hrule height 1.0pt}
PSNR (dB) & 16.25 &  15.71 & 16.21 & \underline{16.69} & 16.02 & \bf17.07 \\
Params (M) & -- & 5 & \underline{0.01} & 20 & \bf0 & 26\\
Time (s) & -- & 1.328 & \bf0.040 & \underline{0.122} & 0.237 & 0.153  \\ 
Offline & -- & $\times$ & $\checkmark$ & $\checkmark$ & $\times$ & $\checkmark$ \\ 
\noalign{\hrule height 1.0pt}
\end{tabular}
\label{Comparison}
\vspace{-0.1in}
\end{table}

\begin{figure}[t!]
    \begin{center}
    \includegraphics[width=1\columnwidth]{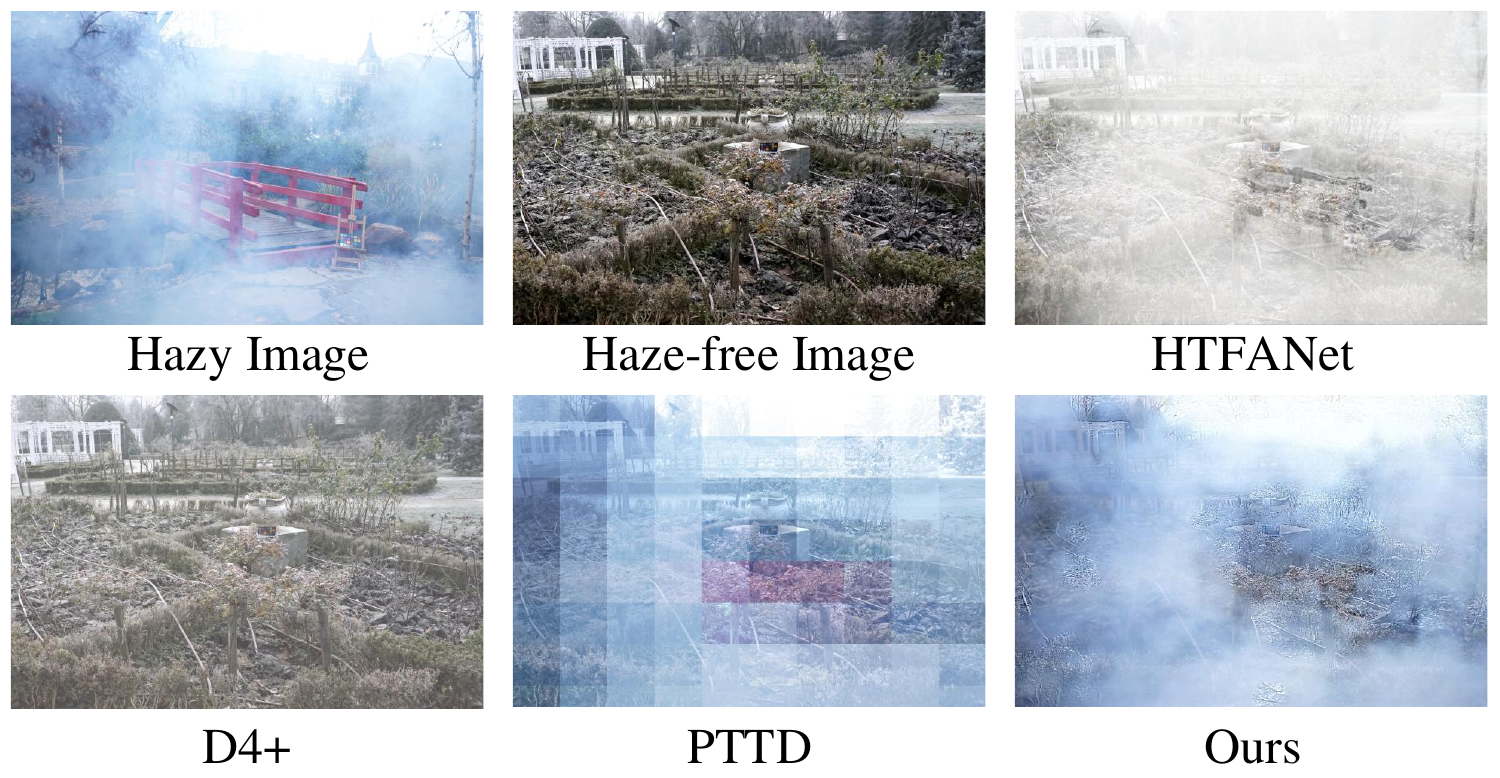}
    \end{center}
    \vspace{-0.2in}
    \caption{Comparison of transferred hazy images among HTFANet~\cite{HTFANet}, D$^4$+~\cite{D4plus}, PTTD~\cite{chen2024promptbased}, and our method.} 
    \label{Haze_Transfer_Comparison_Crop}
    \vspace{-0.1in}
\end{figure} 

\begin{figure}[t!]
    \begin{center}
    \includegraphics[width=1\columnwidth]{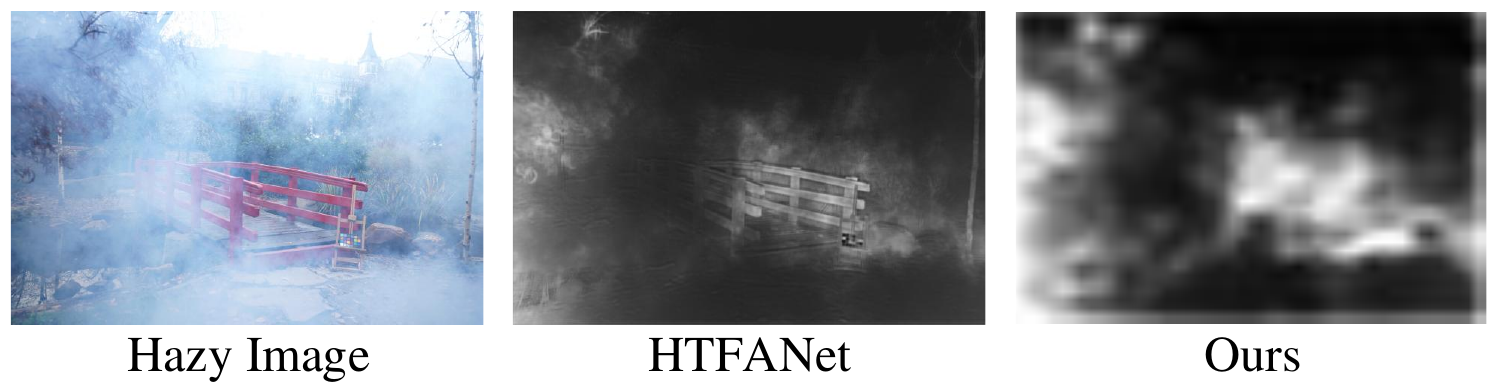}
    \end{center}
    \vspace{-0.2in}
    \caption{Comparison of transmission maps between HTFANet~\cite{HTFANet} and our method.} 
    \label{Transmission_crop}
    \vspace{-0.2in}
\end{figure} 


\section{Conclusion}
We presented an effective domain adaptation framework for image dehazing that leverages a domain-specific fine-tuning set, generated via haze transfer, to adapt dehazing models during testing. At the core of our approach is the Physics-guided Haze Transfer Network (PHATNet), which effectively disentangles and transfers target-domain haze patterns onto source-domain haze-free images. PHATNet incorporates the Atmospheric Scattering Model (ASM) as an inductive bias, enabling efficient separation of haze patterns and scene content within the ASM parametric space. To facilitate accurate haze disentanglement and transfer, we proposed the Haze-Transfer-Consistency Loss to ensure accurate transfer of target-domain haze patterns and the Content-Leakage Loss to prevent scene content leakage from target-domain hazy images. Extensive experiments have demonstrated that PHATNet significantly enhances the performance of state-of-the-art dehazing models across multiple real-world haze datasets, highlighting its effectiveness in domain adaptation for image dehazing.

\paragraph{Acknowledgments} 
This work was supported in part by the National Science and Technology Council (NSTC) under grants 113-2634-F-002-003, 112-2221-E-A49-090-MY3, and 113-2221-E-004-006-MY2. This work was funded in part by MediaTek and NVIDIA Academic Grant.

{
    \small
    \bibliographystyle{ieeenat_fullname}
    \bibliography{main}
}

\end{document}